\documentclass[10pt,twocolumn,letterpaper]{article}

\usepackage{cvpr}
\usepackage{times}
\usepackage{epsfig}
\usepackage{graphicx}
\usepackage{amsmath}
\usepackage{amssymb}
\usepackage{caption}
\usepackage{subcaption}

\newcommand{\mypar}[1]{\vspace{-4mm}\paragraph{#1}}

\DeclareMathAlphabet{\mathpzc}{T1}{pzc}{m}{n}

\newlength{\halfwidth}
\newlength{\fullwidth}
\newlength{\tikzimgheight}
\newlength{\tikzimgwidth}

\usepackage{ifthen}

\usepackage[pagebackref=true,breaklinks=true,letterpaper=true,colorlinks,bookmarks=false]{hyperref}

 \cvprfinalcopy 

\def\cvprPaperID{1259} 

\ifcvprfinal\fi
\begin{document}

\title{We don't need no bounding-boxes: \\
Training object class detectors using only human verification}

\author{\hspace{-0.8cm} Dim P. Papadopoulos \hspace{1.2cm} Jasper R. R. Uijlings \hspace{1.0cm} Frank Keller \hspace{1.7cm} Vittorio Ferrari \\
{\tt\small \hspace{-0.7cm} dim.papadopoulos@ed.ac.uk \hspace{0.4cm} jrr.uijlings@ed.ac.uk \hspace{0.15cm} keller@inf.ed.ac.uk \hspace{0.05cm} vferrari@inf.ed.ac.uk}\\
University of Edinburgh
}

\maketitle

\begin{abstract}

Training object class detectors typically requires a large set of images in which objects are annotated by bounding-boxes. However, manually drawing bounding-boxes is very time consuming.
We propose a new scheme for training object detectors which only requires annotators to verify bounding-boxes produced automatically by the learning algorithm.
Our scheme iterates between re-training the detector, re-localizing objects in the training images, and human verification. We use the verification signal both to improve re-training and to reduce the search space for re-localisation, which makes these steps different to what is normally done in a weakly supervised setting.
Extensive experiments on PASCAL VOC 2007 show that
(1) using human verification
to update detectors and reduce the search space
leads to the rapid production of high-quality bounding-box annotations;
(2) our scheme delivers detectors performing almost as good as those trained in a fully supervised setting, without ever drawing any bounding-box;
(3) as the verification task is very quick, our scheme substantially reduces total annotation time
by a factor $6\times$-$9\times$.

\end{abstract}

\vspace{-0.6cm}
\section{Introduction}

Object class detection is a central problem in computer vision.
Training a detector typically requires a large set of images in which objects have been manually annotated with bounding-boxes~\cite{dalal05cvpr,everingham10ijcv,felzenszwalb10pami,girshick15iccv,girshick14cvpr,MalisiewiczICCV11,uijlings13ijcv,viola:nips05,wang13iccv}.
Bounding-box annotation is tedious, time consuming and expensive. For instance, annotating ILSVRC, currently the most popular object class detection dataset,
required 42s per bounding-box by crowd-sourcing on Mechanical Turk \cite{russakovsky15ijcv} using a technique specifically developed for efficient bounding-box annotation \cite{su12aaai}. 

In order to reduce the cost of bounding-box annotation, researchers have mainly focused on two strategies. 
The first is learning in the weakly supervised setting, i.e. given only labels indicating which object classes are present in an image. 
While this setting is much cheaper, it produces lower quality detectors, typically performing only about half as well as when trained from bounding-boxes~\cite{bilen14bmvc,bilen15cvpr,cinbis15pami,deselaers10eccv,russakovsky12eccv,siva11iccv,song14icml,song14nips,wang15tip}. 
The second strategy is active learning, where the computer requests human annotators to draw bounding-boxes on a subset of images actively selected by the learner itself. 
This strategy can produce high quality detectors, but it still requires humans to draw a lot of
bounding-boxes in order to get there, leading to limited gains in terms of total annotation
time~\cite{vijayanarasimhan14ijcv,yao2012cvpr}. 


In this paper we propose a new scheme for learning object detectors which only requires humans to {\em verify} bounding-boxes produced automatically by the learning algorithm:
the annotator merely needs to decide whether a bounding-box is correct or not.
Crucially, answering this verification question takes much less time than actually drawing the bounding-box. 

Given a set of training images with image-level labels, our scheme 
iteratively alternates between updating object detectors,
re-localizing objects in the training images,
and querying humans for verification.
At each iteration we use the verification signal in two ways.
First, we update the object class detector using only positively verified bounding-boxes.
This makes it stronger than when using all detected bounding-boxes, as it is commonly done in the weakly supervised setting, because typically many of them are incorrect. 
Moreover, once the object location in an image has been positively verified, it can be fixed and removed from consideration 
in subsequent iterations. 
Second, we observe that bounding-boxes judged as incorrect still provide valuable information about where the object is not. Building on this observation, we use the negatively verified bounding-boxes to reduce the search space of possible object locations in subsequent iterations.
Both these points help to rapidly find more objects in the remaining images. 
This results in a framework for training object detectors which minimizes human
annotation effort and eliminates the need to draw \emph{any} bounding-box.

\begin{figure*}[t]
    \vspace{-.7cm}\center
\includegraphics[width=1\linewidth]{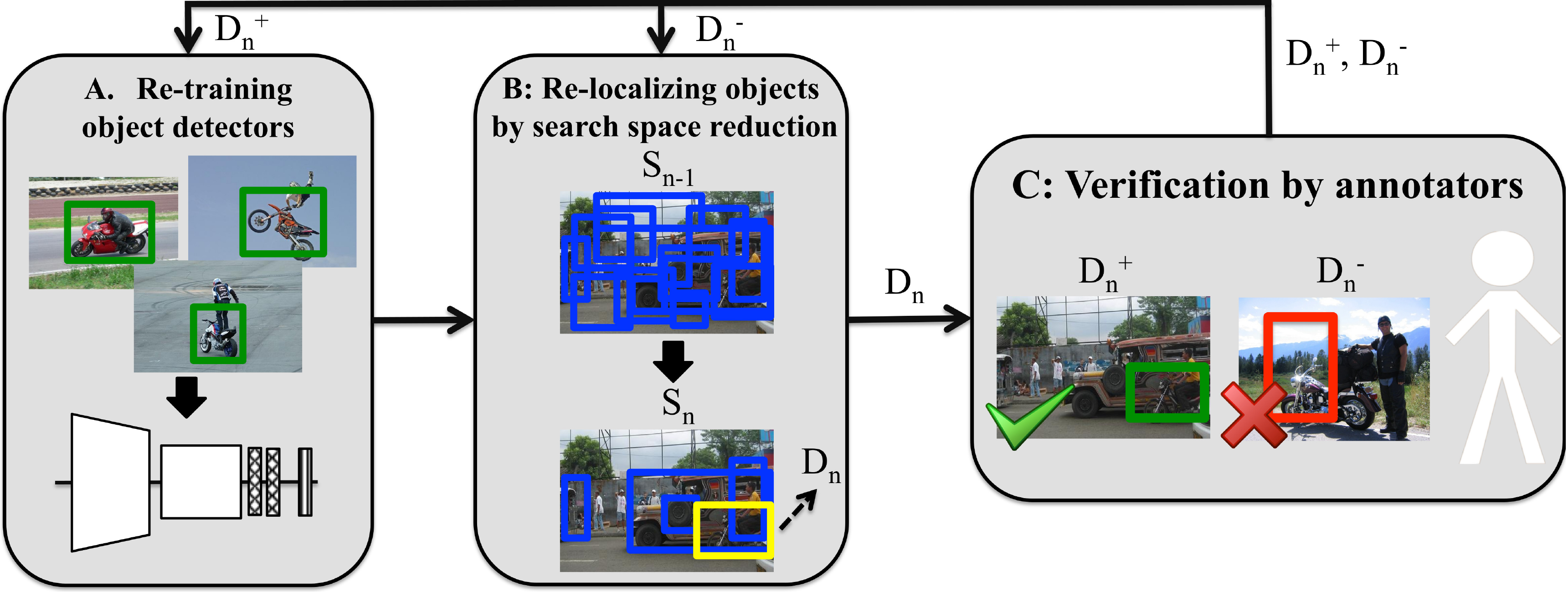}
\vspace{-.4cm}
\caption{\small Our framework iterates between (A) re-training object detectors, (B) re-localising objects,
and (C) querying annotators for verification. The verification signal resulting from (C) is used in
both (A) and (B).}
\label{fig:Scheme2}
\vspace{-.4cm}
\end{figure*}




In extensive experiments on the popular PASCAL VOC 2007 dataset~\cite{everingham10ijcv} with both simulated and actual annotators, we show that:
(1) using human verification to update detectors and reduce the search space leads to rapid
production of high-quality bounding-box annotations;
(2) our scheme delivers object class detectors performing almost as good as those trained in a fully supervised setting, {\em without ever drawing any bounding-box};
(3) as the verification task is very quick, our scheme substantially reduces total annotation time
by a factor $6\times$-$9\times$. 


\section{Related Work}


\paragraph{Weakly-supervised object localization (WSOL).}
Many previous techniques~\cite{bilen14bmvc,bilen15cvpr,cinbis14cvpr,deselaers10eccv,russakovsky12eccv,siva11iccv,song14icml,song14nips,wang15tip} try to learn object class detectors in the weakly supervised setting,
i.e. given training images known to contain instances of a certain object class, but not their
location. The task is both to localize the objects in the training images and to learn an object 
detector for localizing instances in new images.

Recent work on WSOL~\cite{bilen14bmvc,bilen15cvpr,cinbis14cvpr,song14icml,song14nips,wang15tip} has shown
remarkable progress, mainly because of the use of Convolutional Neural Nets
(CNN~\cite{girshick14cvpr,krizhevsky12nips}), which greatly improve visual recognition tasks.
However, learning a detector without location annotation is difficult and performance is
still quite low compared to fully supervised methods (typically about half the mAP of the same detection model trained on the same images but with manual bounding-box annotation~\cite{bilen14bmvc,bilen15cvpr,cinbis14cvpr,deselaers10eccv,russakovsky12eccv,siva11iccv,song14icml,song14nips,wang15tip}).

Often WSOL is conceptualised as Multiple Instance Learning
(MIL)~\cite{bilen14bmvc,cinbis14cvpr,deselaers10eccv,dietterich97ai,shi12bmvc,siva11iccv,song14icml,song14nips}.
Images are treated as bags of windows (instances). A negative image contains only negative
instances.  A positive image contains at least one positive instance, mixed in with a majority of
negative ones.  The goal is to find the true positive instances from which to learn a window
classifier for the object class. This typically happens by iteratively alternating between
(A) re-training the object detector given the current selection of positive instances and
(B) re-localising instances in the positive images using the current object detector.
Our proposed scheme also contains these steps. However, we introduce a human verification step, whose signal is fed into both (A) and (B), which fundamentally alters these steps. The resulting framework leads to
substantially better object detectors with modest additional annotation effort
(sec.~\ref{sec:pascal_human}).

\mypar{Humans in the loop.}
Human-machine collaboration approaches have been successfully used in tasks that are currently too difficult to be solved by computer vision alone, such as fine-grained visual
recognition~\cite{branson10eccv,deng13cvpr,wah11iccv,wah14cvpr}, semi-supervised
clustering~\cite{lad14eccv}, attribute-based image
classification~\cite{biswas13cvpr,parikh11iccv,parkash12eccv}.
These works combine the responses of pre-trained computer vision models on a new test image with human input to fully solve the task.
In the domain of object detection,
Russakovsky et al.~\cite{russakovsky15cvpr} propose such a scheme to fully detect all objects in images of complex scenes.
Importantly, their object detectors are pre-trained on bounding-boxes from the large training set of ILSVRC 2014~\cite{russakovsky15ijcv}, as their goal is not to make an efficient training scheme.

\mypar{Active learning.}
Active learning schemes iteratively train models while requesting humans to annotate 
a subset of the data points actively selected by the learner as being the most informative.
Previous active learning work has mainly focussed on image
classification~\cite{joshi09cvpr,kapoor07iccv,kovashka11iccv,qi08cvpr}, and free-form region
labelling~\cite{siddiquie10cvpr,vijayanarasimhan08nips,vijayanarasimhan09cvpr}.

A few works have proposed active learning schemes specifically for training object class detectors~\cite{vijayanarasimhan14ijcv,yao2012cvpr}.
Vijayanarasimhan and Grauman~\cite{vijayanarasimhan14ijcv} propose an approach where the training images do not come from a predefined dataset but are crawled from the web.
Here annotators are asked to draw many bounding-boxes around the target objects
(about one third of the training images~\cite{vijayanarasimhan14ijcv}).
Yao et al.~\cite{yao2012cvpr} propose to manually correct bounding-boxes detected in video. While
both~\cite{vijayanarasimhan14ijcv,yao2012cvpr} 
produce high quality detectors, they achieve only moderate gains in annotation time,
because drawing or correcting bounding-boxes is expensive.
In contrast, our scheme only asks annotators to verify bounding-boxes, never to draw. This leads to more substantial reductions in annotation time.




\mypar{Other ways to reduce annotation effort.}
A few authors tried to learn object detectors from videos, where the spatio-temporal coherence of the video frames facilitates object localization~\cite{Leistner11,prest12cvpr,Tang2013}. An alternative is transfer learning, where learning a model for a new class is helped by labeled examples of related classes~\cite{Aytar11iccv,fei2007CVIU,guillaumin12cvpr,kuettel12eccv,lampert:cvpr09}. Hoffman et al.~\cite{hoffman14nips} proposed an algorithm that transforms an image classifier to an object detector without bounding-box annotated data using domain adaptation. Other types of data, such as text from web pages or newspapers~\cite{Berg04a,DuyguluECCV02,Gupta08:eccv,luo09nips} or eye-tracking data~\cite{papadopoulos14eccv}, have also been used as a weak annotation signal to train object detectors.


\section{Method}

In this paper we are given a training set with image-level labels. Our goal is to obtain 
object instances annotated by bounding-boxes and to train good object detectors while minimizing human annotation effort. We
therefore propose a framework where annotators only need to \emph{verify} bounding-boxes automatically produced by our scheme.

Our framework iteratively alternates between (A) re-training object detectors, (B) re-localizing
objects in the training images, and (C) querying annotators for verification (fig.~\ref{fig:Scheme2}).
Importantly, we use verification signals to help both re-training and re-localisation.


More formally, let $I_n$ be the set of images for which we do not have positively verified bounding-boxes at
iteration $n$ yet. Let $S_n$ be the corresponding set of possible object locations. Initially, $I_0$
is the complete training set and $S_0$ is a complete set of object
proposals~\cite{alexe10cvpr,dollar14eccv,uijlings13ijcv} extracted from these images (we use
EdgeBoxes~\cite{dollar14eccv}).
To facilitate exposition, we describe our framework starting from the verification step (C, sec~\ref{cptVerification}).
At iteration $n$ we have a set of automatically detected bounding-boxes $D_n$ which are
given to annotators to be verified. Detections which are judged to be correct $D_n^+ \subseteq D_n$
are used for re-training the object detectors (A) in the next iteration (sec.~\ref{cptRetraining}).
The verification signal is also used to reduce the search space $S_{n+1}$ for re-localisation (B, sec.~\ref{cptLocalizing}).
We describe our main three steps below. We defer to Sec.~\ref{cptImplementation} a description of the object detection model we use, and of how to automatically obtain initial detections $D_0$ to start the process.


\subsection{Verification by annotators}
\label{cptVerification}

\begin{figure}[t]
\includegraphics[width=\linewidth]{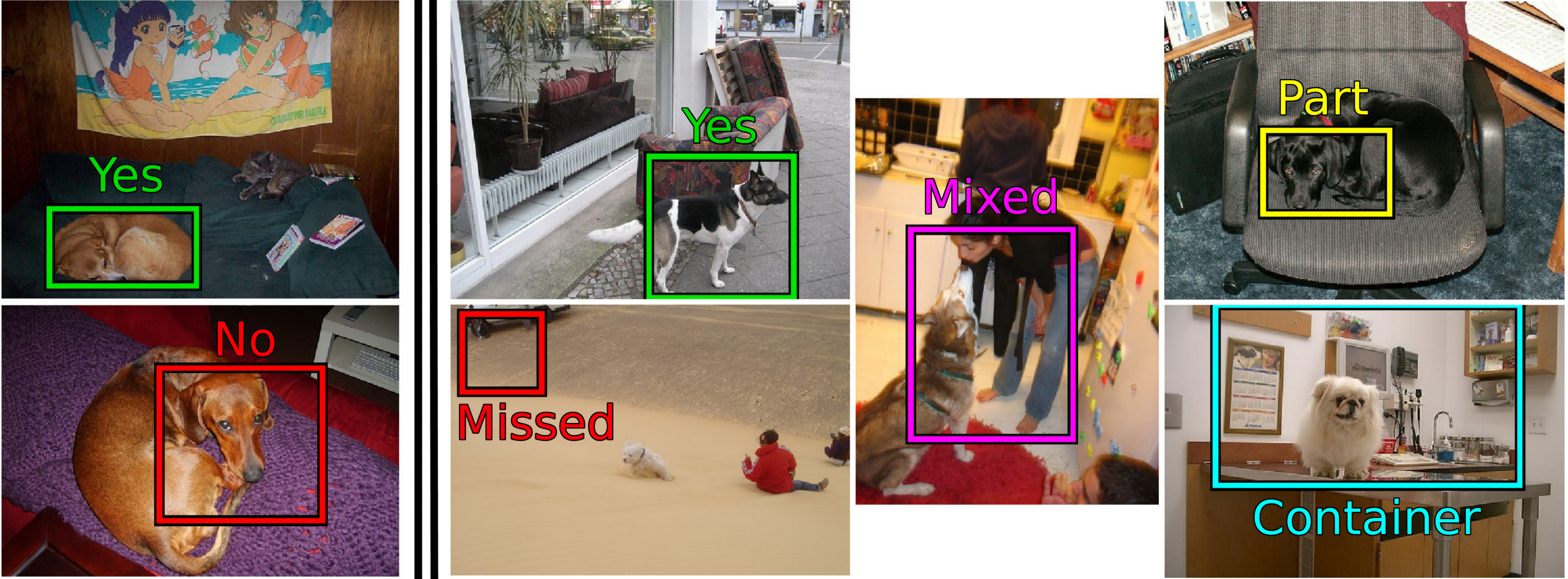}
\vspace{-7mm}
\caption{\small Our two verification strategies for some images of the dog class.
Yes/No verification (left): verify a detection as either correct (\textcolor{green}{Yes}) or incorrect (\textcolor{red}{No}).
YPCMM verification (right): label a detection as \textcolor{green}{Yes}, \textcolor{yellow}{Part}, \textcolor{cyan}{Container}, \textcolor{magenta}{Mixed} or \textcolor{red}{Missed}.}
\label{fig:ExamplesPCMYesNo}
\vspace{-4mm}
\end{figure}

In this phase, we ask annotators to verify the automatically generated detections $D_n$ at
iteration $n$. For this we explore two strategies~(fig.~\ref{fig:ExamplesPCMYesNo}): simple yes/no
verification, and more elaborate verification in which annotators are asked to categorize the type of
error. 

\mypar{Yes/No Verification.} 

In this task the annotators are shown a detection $l_d$ and a class label. They are
instructed to respond Yes if the detection correctly localizes an object of that class, and No
otherwise. This splits the set of object detections $D_n$ into $D_n^+$ and $D_n^-$.
We define ``correct localization'' based on the standard PASCAL Intersection-over-Union 
criterion~\cite{everingham10ijcv} (IoU). Let $l_d$ be the detected object bounding-box and
$l_{gt}$ be the actual object bounding-box (which is not given to the annotator). 
Let $\mathrm{IoU}(l_a, l_b) = | l_a \cap l_b | / | l_a \cup l_b |$, where $|\cdot|$ denotes area.
If $\mathrm{IoU}(l_{gt}, l_d) \geq 0.5$, the detected bounding-box should be considered correct and the annotator should answer Yes.
Intuitively, this Yes/No verification is a relatively simple task which should translate into fast annotation
times.

\mypar{Yes/Part/Container/Mixed/Missed Verification.}

In this task, we asked the annotators to label an object detection $l_d$ as Yes (correct), Part,
Container, Mixed, or Missed. Yes is defined as above ($\mathrm{IoU}(l_{gt}, l_d) \geq 0.5$).
For incorrect detections the annotators are asked to diagnose the error as either
\emph{Part} if it contains part of the target object and no background;
~\emph{Container} if it contains the whole object and some background;
~\emph{Mixed} if it contains part of the object and some background;
~\emph{Missed} if the object was completely missed.
This verification step splits $D_n$ into $D_n^+$ and $D_n^{ypcmm-}$.
Intuitively, determining the type of error is more difficult leading to longer annotation times, but
also brings more information that we can use in the next steps.


\subsection{Re-training object detectors}
\label{cptRetraining}

In this step we re-train object detectors. After the verification step we know that $D_n^+$
contains well localized object instances, while $D_n^-$ or $D_n^{ypcmm-}$ do not. Hence we train using only
bounding-boxes $D_1^+ \cup \cdots \cup D_n^+$ that have been positively verified in past iterations.
To obtain background training samples, we sample proposals which have an IoU in range $[0-0.5)$ with positively verified bounding-boxes.

Note how it is common in WSOL works~\cite{bilen14bmvc,bilen15cvpr,cinbis14cvpr,deselaers10eccv,russakovsky12eccv,siva11iccv,song14icml,song14nips,wang15tip} to also have a re-training step.
However, they typically use all detected bounding-boxes $D_n$. Since in WSOL generally less than half of them are correct, this leads to rather weak detectors.
In contrast, our verification step enables us to train purely from correct bounding-boxes, resulting in stronger, more reliable object detectors. 

\subsection{Re-localizing objects by search space reduction}
\label{cptLocalizing}

In this step we re-localize objects in the training images. For each image, we apply the current object detector to score the object proposals in it, and select the proposal with the highest score as the new detection for that image. 
Importantly, we do not evaluate {\em all} proposals $S_0$, but instead use the verification signal to reduce the search space by removing proposals.

Positively verified detections $D_n^+$ are correct by definition and therefore their images need not be considered in subsequent iterations, neither in the re-localization step nor in the verification step.
For negatively verified detections we reduce the search space depending on the verification strategy, as described below.

\mypar{Yes/No Verification.}
In the case where the annotator judges a detection as incorrect ($D_n^-$), we can simply eliminate its proposal from the search space. This results in the updated search space $S_{n+1}$, where one proposal has been removed from each image with an incorrect detection. 

However, we might make a better use of the negative verification signal.
Since an incorrect detection has an IoU $< 0.5$ with the true bounding-box, we can eliminate all
proposals with an IoU $\geq 0.5$ with it.
This is a more aggressive reduction of the search space.
While it may remove some proposals which are correct according to the IoU criterion, it will not remove the best possible proposal.
Importantly, this strategy eliminates those areas of the search space that matter: high scoring
locations which are unlikely to contain the object. In Sec.~\ref{sec:pascal_sim} we
investigate which way of using negatively verified detection performs better in practice.

\mypar{Yes/Part/Container/Mixed/Missed Verification.}

\begin{figure}[t]
    \vspace{-.5cm}
\includegraphics[width=\linewidth]{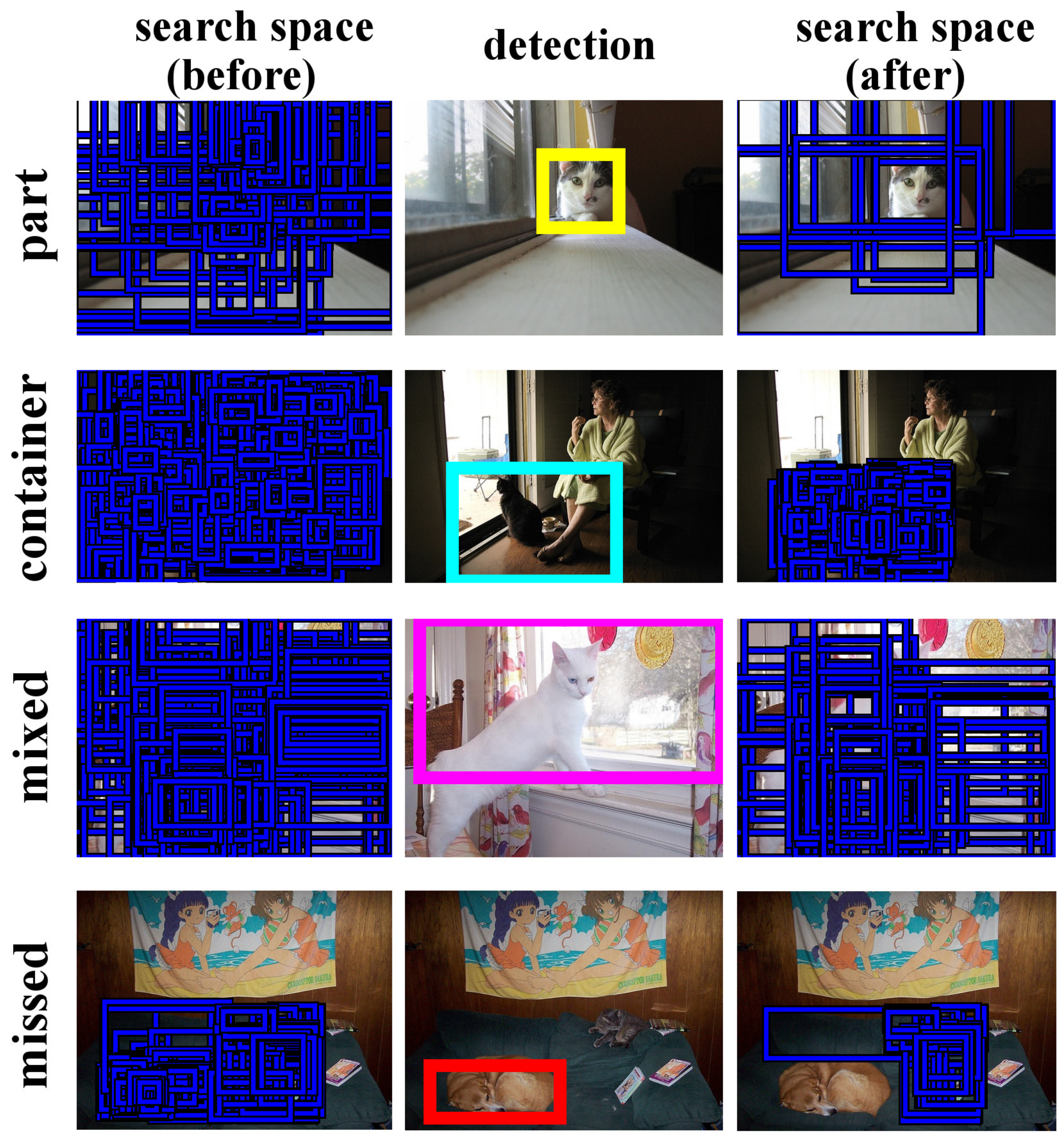}
\vspace{-7mm}
\caption{\small Visualisation of \textcolor{blue}{search space} reduction induced by YPCMM verification on some images of the cat class (\textcolor{yellow}{part}, \textcolor{cyan}{container}, \textcolor{magenta}{mixed}, and \textcolor{red}{missed}). In the last row, the search space reduction steers the re-localization process towards the small cat on the right of the image and away from the dog on the left.}
    \label{fig:ReduceSpace}
    \vspace{-4mm}
\end{figure}

In the case where annotators categorize incorrect detections as Part/Container/Mixed/Missed, we can use the type of error to get an even greater reduction of the search space.
Depending on the type of error we eliminate different proposals (fig.~\ref{fig:ReduceSpace}):
{\em Part:} eliminate all proposals which do not contain the detection;
{\em Container:} eliminate all proposals which are not inside the detection;
{\em Mixed:} eliminate all proposals which are not inside the detection, or do not contain it, or have zero IoU with it, or have IoU $\geq 0.5$ with it;
{\em Missed:} eliminate all proposals which have non-zero IoU with the detection.

To precisely determine what is ``inside'' and ``contained'', we introduce the Intersection-over-A measure:
$\mathrm{IoA}(l_a, l_b) = | l_a \cap l_b | / |l_a|$.
Note that $\mathrm{IoA}(l_{gt},l_d) = 1$ if the detection $l_d$ contains the true object bounding-box $l_{gt}$, whereas $\mathrm{IoA}(l_d,l_{gt}) = 1$ if $l_d$ covers a part of $l_{gt}$.
In practice, if $l_d$ is judged to be a {\em Part} by the annotator, we eliminate all proposals $l_s$ with $\mathrm{IoA}(l_d,l_s) \leq 0.9$.
Similarly, if  $l_d$ is judged to be a {\em Container}, we eliminate all proposals $l_s$ with $\mathrm{IoA}(l_s,l_d) \leq 0.9$. We keep the tolerance threshold $0.9$ fixed in all experiments.

%

Note how in WSOL there is also a re-localisation step. However, because there is no verification signal there
is also no search space reduction: each iteration needs to consider the complete set $S_0$ of proposals. In contrast, in our work the search space reduction greatly facilitates re-localization.

\section{Implementation details}
\label{cptImplementation}

We summarize here two existing state-of-the-art components that we use in our framework:
the object detection model, and a WSOL algorithm which we use to obtain initial object detections $D_0$.

\begin{figure*}[t]
    \vspace{-.4cm}\center
\includegraphics[width=1\linewidth]{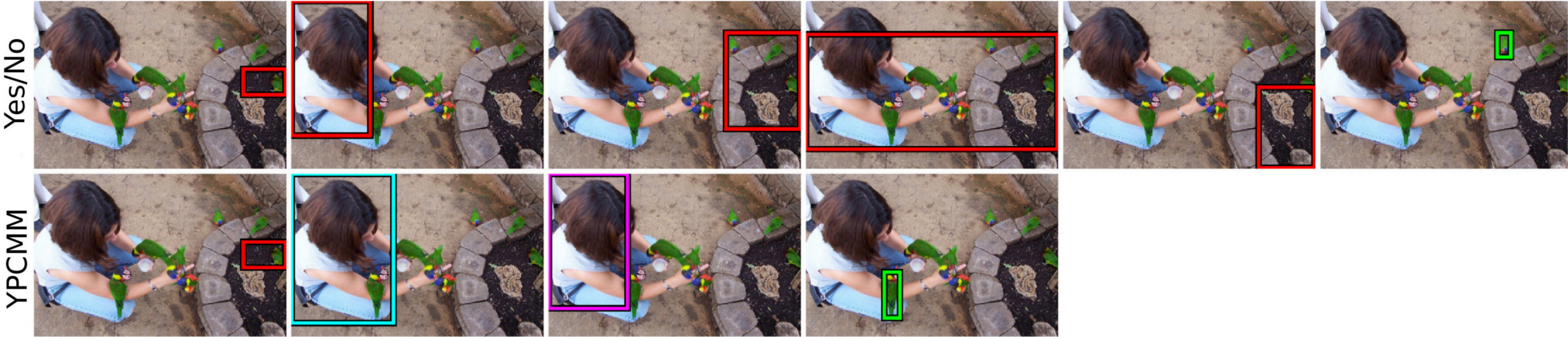}
\vspace{-7mm}
\caption{\small Comparing the search process of Yes/No verification with Yes/Part/Container/Mixed/Missed on an image of the bird class.
The extra signal for YPCMM allows for a more targeted search, resulting in fewer verification steps
to find the object.}
\vspace{-3mm}
\label{fig:Examples}
\end{figure*}

\subsection{Object class detector}
\label{cptFastRCNN}

As object detector we use Fast R-CNN~\cite{girshick15iccv}, which combines
object proposals~\cite{alexe10cvpr,dollar14eccv,uijlings13ijcv} with
CNNs~\cite{he14eccv,krizhevsky12nips,simonyan15iclr}.
%
Instead of Selective Search~\cite{uijlings13ijcv} we use EdgeBoxes~\cite{dollar14eccv} which gives us an ``objectness''
measure~\cite{alexe10cvpr} which we use in the initialization phase described below.
%
For simplicity of implementation, for the re-training step (sec.~\ref{cptRetraining}) we omit bounding-box regression, so that the set of object proposals stays fixed throughout all iterations. For evaluation on the test set, we then train detectors with bounding-box regression.
In most of our experiments, we use AlexNet~\cite{krizhevsky12nips} as the underlying CNN architecture.

%


\subsection{Initialization by Multiple Instance Learning}
\label{cptMIL}

We perform multiple instance learning (MIL) for weakly supervised object
localisation~\cite{bilen14bmvc,cinbis14cvpr,song14icml} to obtain the initial set of detections
$D_0$. We start with the training images $I_0$ and the set of object proposals $S_0$ extracted using
EdgeBoxes~\cite{dollar14eccv}.
Following~\cite{bilen14bmvc,girshick14cvpr,song14icml,song14nips,wang15tip} we extract CNN features on
top of which we train an SVM. We iterate between (A) re-training object detectors and (B)
re-localizing objects in the training images. We stop when two subsequent re-localization steps
yield
the same detections, which typically happens within 10 iterations. These detections
become $D_0$. In the very first iteration, we train the
classifier using complete images as positive training examples~\cite{cinbis14cvpr,russakovsky12eccv}. 

We apply two improvements to the standard MIL framework.
First, in high dimensional feature space the discriminative SVM classifier can relatively easily separate any positive examples from negative examples, which means that most positive examples are far from the decision hyperplane. Hence the same positive training examples used for re-training (A) are often re-localised in (B), leading to premature locked-in behaviour. To prevent this Cinbis et al.~\cite{cinbis14cvpr,cinbis15pami} introduced
multi-fold MIL: similar to cross-validation, the dataset is split into $10$ subsets, where the
re-localisation on each subset is done using detectors trained on the union of all other subsets.
Second, like in~\cite{cinbis15pami,deselaers10eccv,guillaumin12cvpr,prest12cvpr,shapovalova12eccv,siva11iccv,shi12bmvc,tang14cvpr,wang14eccv-cosegmentation}, we combine the object detector score with a general measure of ``objectness''~\cite{alexe10cvpr}, which measures how likely it is that a proposal tightly encloses an object of any class (e.g. bird, car, sheep), as opposed to background (e.g. sky, water, grass). In this paper we use the recent objectness measure of~\cite{dollar14eccv}.

\section{Experimental Results}


\subsection{Dataset and evaluation protocol}

\paragraph{PASCAL VOC 2007.}
We perform expriments on PASCAL VOC 2007~\cite{everingham10ijcv}, which consists of 20 classes. The trainval set contains 5011 images, while the test set contains 4952 images.
We use the trainval set with accompanying image-level labels to train object detectors, and measure their performance on the test set.
Following the common protocol for WSOL experiments~\cite{cinbis14cvpr,cinbis15pami,deselaers10eccv,russakovsky12eccv,wang15tip}, we exclude trainval images that contain only difficult and truncated instances, ending up with 3550 images.
In sections~\ref{sec:pascal_sim}, \ref{sec:pascal_human} we carry out a detailed analysis of our system in these settings, using AlexeNet as CNN architecture~\cite{krizhevsky12nips}.
For completeness, in section~\ref{sec:pascal_full} we also present results when using the complete trainval set and the deeper VGG16 as CNN architecture~\cite{simonyan15iclr}.


\mypar{Evaluation.}
Given a training set with image-level labels, our goal is to localize the object
instances in this set and to train good object detectors, while minimizing human annotation
effort. We evaluate this by exploring the trade-off between localization performance and quality of
the object detectors versus required annotation effort.
We quantify localization performance in the training set with the Correct Localization (CorLoc)
measure~\cite{bilen14bmvc,bilen15cvpr,cinbis14cvpr,cinbis15pami,deselaers10eccv,russakovsky12eccv,siva11iccv,wang15tip}.
CorLoc is the percentage of images in which the bounding-box returned by the algorithm correctly localizes an object of the target class (i.e., IoU $\geq 0.5$).

We quantify object detection performance on the test set using mean average precision
(mAP), as standard in
PASCAL VOC 07.
We quantify annotation effort both in terms of the number of verifications and in terms of actual human time measurements. 

As most previous WSOL methods~\cite{bilen14bmvc,bilen15cvpr,cinbis14cvpr,cinbis15pami,deselaers10eccv,russakovsky12eccv,siva11iccv,song14icml,song14nips,wang15tip}, our scheme returns exactly one bounding-box per class per training image. This enables clean comparisons to previous work in terms of CorLoc on the training set, and keeps the human verification tasks simple (as we do not need to ask the annotators whether they see additional instances in an image).
Note how at test time the detector is capable of localizing multiple objects of the same class in the same image (and this is captured in the mAP measure).

\begin{figure}[t!]
\vspace{-.4cm}
\includegraphics[width=\linewidth]{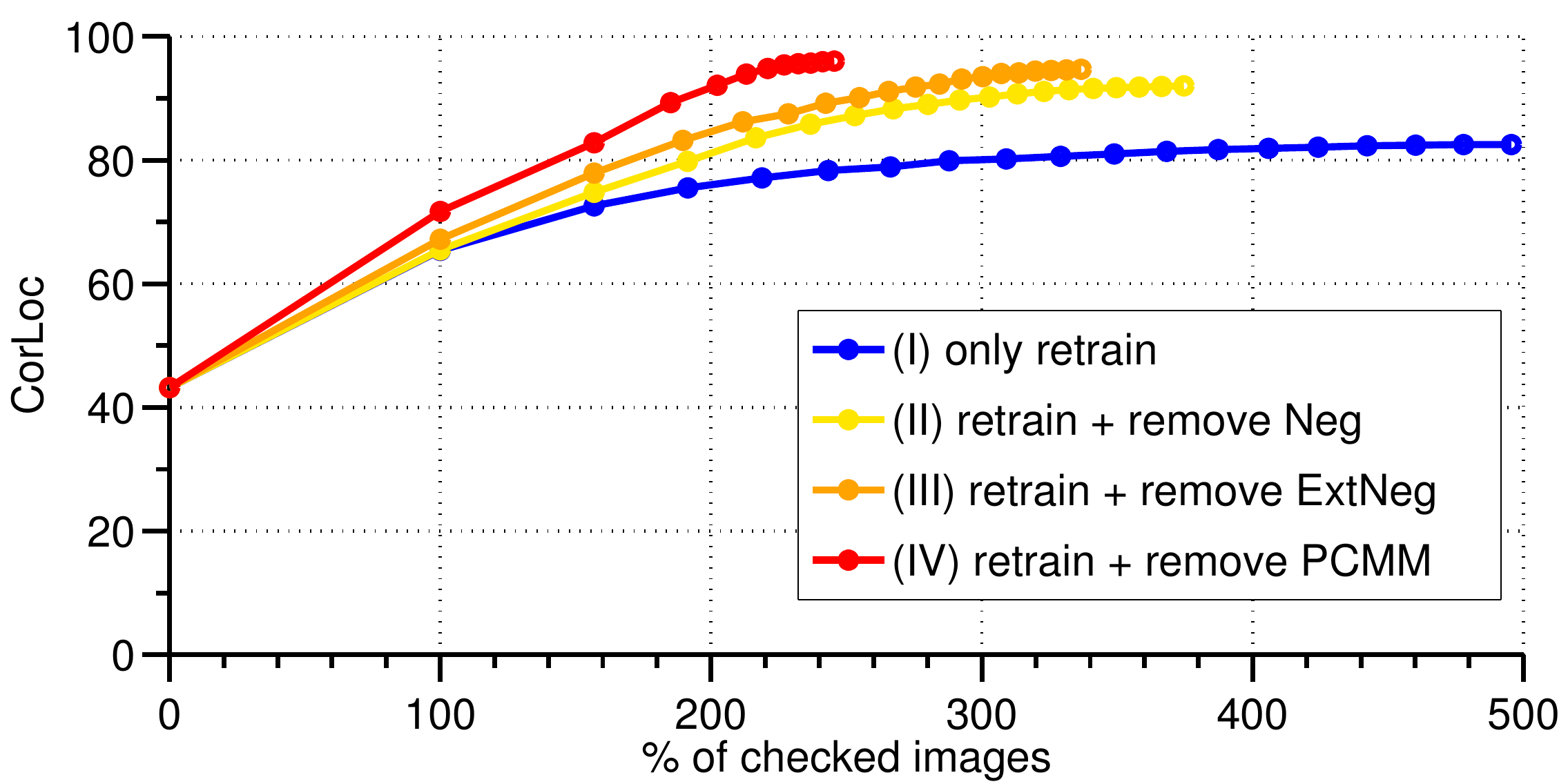}
\vspace{-7mm}
\caption{\small Trade-off between the number of verifications and CorLoc for the simulated verification case on PASCAL VOC 2007.}
\label{fig:Corloc_sim}
\vspace{-4mm}
\end{figure}


\mypar{Compared methods.}
We compare our approach to the fully supervised alternative by training the same object detector (sec.~\ref{cptFastRCNN}) on the same training images, but with manual bounding-boxes (again, one bounding-box per class per image). 
On the other end of the supervision spectrum, we also compare to a modern MIL-based WSOL technique (sec~\ref{cptMIL}) run on the same training images, but without human verification. Since that technique also forms the initialization step of our method, this comparison reveals how much farther we can go with human verification.

For MIL WSOL, the effort to draw bounding-boxes is zero. For fully supervised learning we take the actual annotation times for ILSVRC from~\cite{su12aaai}: they report 26 seconds for drawing one bounding-box without quality control, and 42 seconds with quality control.
These timings are also representative for PASCAL VOC, since it is of comparable difficulty and its annotations are of comparable quality, as discussed in~\cite{russakovsky15ijcv}.
The bounding-boxes in both datasets are of high quality and precisely match the object extent. 



\subsection{Simulated verification}
\label{sec:pascal_sim}

We first use simulated verification to determine how best to use
the verification signal. We simulate human verification by using the available ground-truth bounding
boxes. Note how these are {\em not} given to the learning algorithm, they are only used to derive the verification signals of sec.~\ref{cptVerification}.
Fig.~\ref{fig:Corloc_sim} compares four ways to use the verification signal in terms of the
trade-off between the number of verifications and CorLoc (sec.~\ref{cptLocalizing}): 
\textbf{(I) only retrain} the object detector (using positively verified detections $D_n^+$);
\textbf{(II) retrain + remove Neg}: for Yes/No verification, retrain and reduce the search space by eliminating one proposal for each negatively verified detection;
\textbf{(III) retrain + remove ExtNeg}: for Yes/No verification, retrain and eliminate all proposals overlapping with a negatively verified detection;
\textbf{(IV) retrain + remove PCMM}: for YPCMM verification, retrain and eliminate proposals according to the type of error.

As fig.~\ref{fig:Corloc_sim} shows, even using verification just to re-train the object detector (I)
drastically boosts CorLoc from the initial 43\% (achieved by MIL WSOL) up to 82\%.
This requires checking each training image on average 4 times.
Using the verification signal in the re-localisation step by reducing the search space (II--IV) helps to reach this CorLoc substantially faster (1.6--2 checks per image).
Moreover, the final CorLoc is much higher when we reduce the search space.
Removing negatively verified detections brings a considerable jump to 92\% CorLoc (II);
removing all proposals around negatively verified detections further increases CorLoc to 95\% (III);
while the Yes/Part/Container/Mixed/Missed strategy appears to be the most promising, achieving a near-perfect 96\% CorLoc after checking each image only 2.5 times on average.
These results show that feeding the verification signal into both the re-training and
re-localisation steps quickly results in a large number of correctly located object instances.


Fig.~\ref{fig:Examples} compares the search process of Yes/No and YPCMM verification strategies on
a bird example. The second detection is diagnosed in YPCMM as a container. This focuses the search
to that particular part of the image. In contrast, detections of the Yes/No case jump around before
finding the detection. This shows that the search process of YPCMM is more targeted. However, in
both cases the target object location is found rather quickly.

In conclusion, YPCMM is the most promising verification strategy, followed by Yes/No with removing all proposals overlapping with a negatively verified detection (III).
Since Yes/No verification is intuitively easier and faster, we try both strategies in 
experiments with human annotators.

\subsection{Human verification}
\label{sec:pascal_human}

\begin{figure}[t]
\vspace{-4mm}
\includegraphics[width=\linewidth]{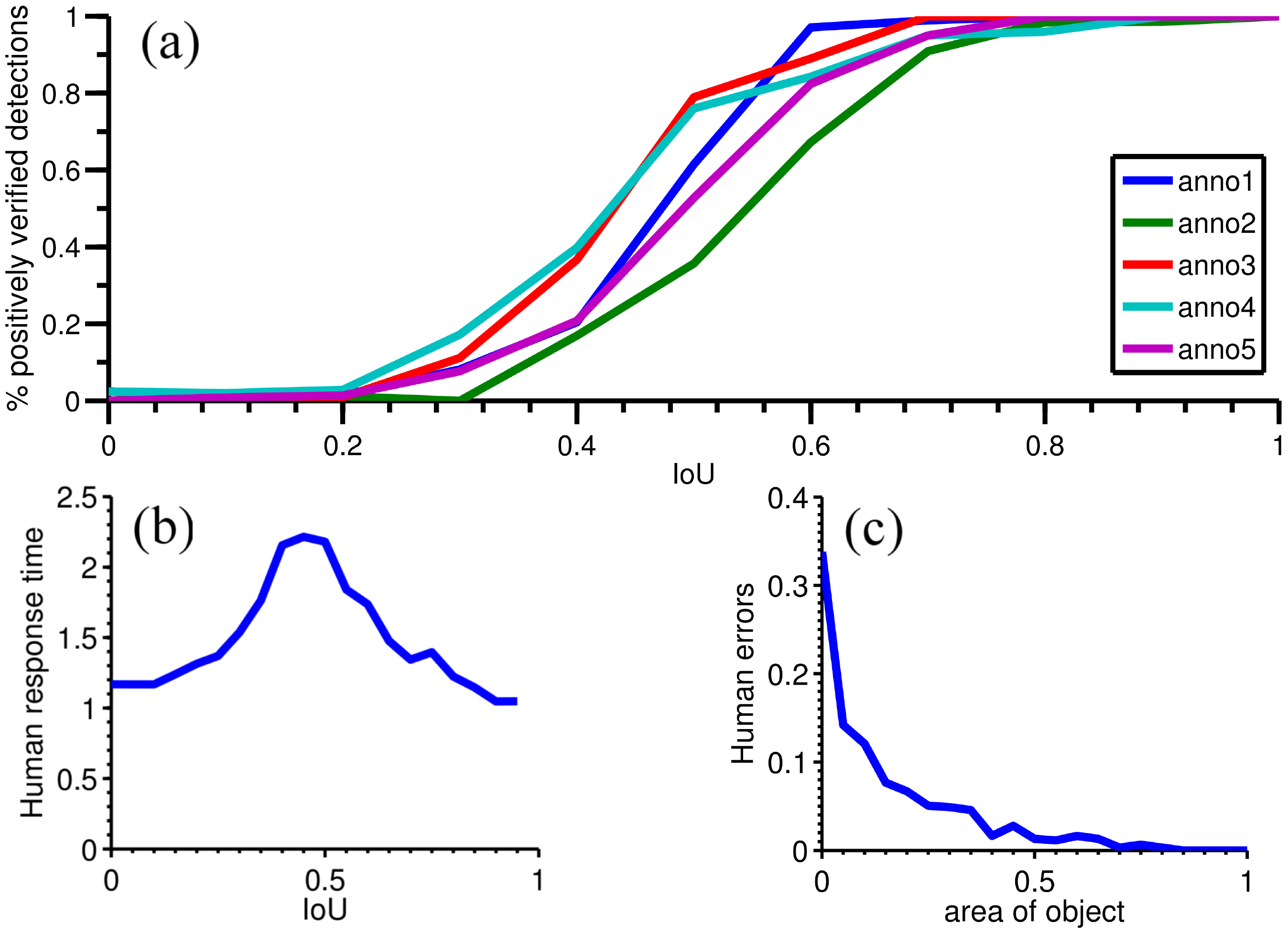}
\vspace{-7mm}
\caption{\small (a) Percentage of positively verified detections as a function of ground-truth IoU, for each annotator. (b) Average human response time as a function of ground-truth IoU.
(c) Percentage of incorrectly verified detections as a function of object area (relative to image area).}
\label{fig:HumanAnalysis}
\vspace{-4mm}
\end{figure}

\begin{figure*}[t!]
\vspace{-4mm}
\includegraphics[width=\linewidth]{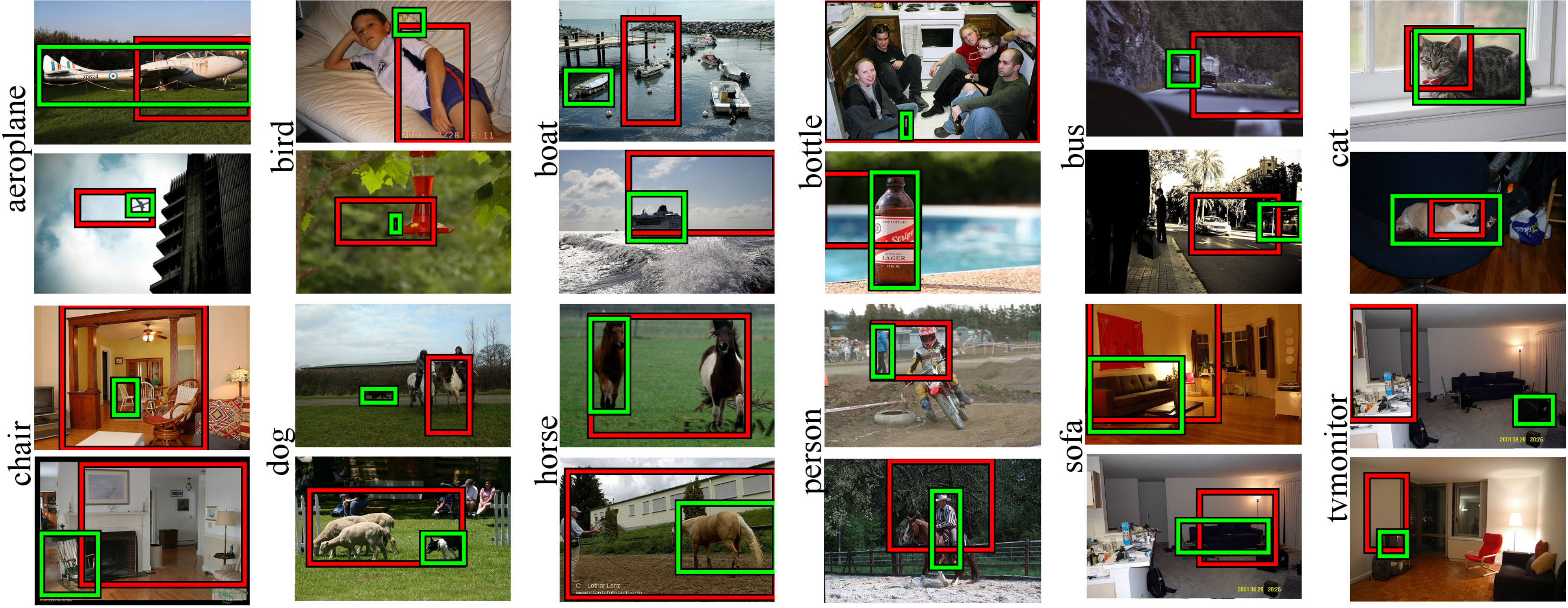}
\vspace{-7mm}
\caption{\small Examples of objects localized by using our proposed Yes/No human verification scheme on the trainval set of PASCAL VOC 2007 (sec.~\ref{sec:pascal_human}). For each example, we compare the final output of our scheme (green box) to the output of the reference multiple instance learning (MIL) weakly-supervised object localization approach (red box) (sec.~\ref{cptMIL}).}
\label{fig:WS_to_verification}
\vspace{-4mm}
\end{figure*}


\vspace{4mm}
\mypar{Instructions and interface.} For the Yes/No and YPCMM verification tasks, we used five annotators from our university who
were given examples to learn about the IoU criterion.
For both tasks we created a full-screen interface. All images of a target class were shown in sequence, with the current detection superimposed.
For Yes/No verification, annotators were asked to press ``1'' for Yes and ``0''
for No. This simple task took on average $1.6$ seconds per verification. 
For YPCMM verification, the annotators were asked to click on one of five on-screen buttons corresponding
to Yes, Part, Container, Mixed and Missed. This more elaborate task
took on average $2.4$ seconds per verification.

\mypar{Analysis of human verification.}



Fig.~\ref{fig:HumanAnalysis}a reports the percentage of positively verified detections as a function of their IoU with the ground-truth bounding-box. We observe that humans behave quite closely to the desired PASCAL criterion (i.e. IoU $>0.5$) which we use in our simulations. All annotators behave identically on easy cases (IoU $< 0.25$, IoU $> 0.75$). On boundary cases (IoU $\approx 0.5$) we observe some annotator bias.
For example, $anno2$ tends to judge boundary cases as wrong detections, whereas $anno3$ and $anno4$ judge them more frequently as correct. Overall the percentage of incorrect Yes and No judgements are 14.8\% and 8.5\%, respectively. Therefore there is a slight bias towards Yes, i.e. humans tend to be slightly more lenient than the IoU$> 0.5$ criterion.

While the average human response time is $1.6$ s for the Yes/No verification, the response time for verifying difficult detections (IoU $\approx 0.5$) is significantly higher ($2.2$ s, fig.~\ref{fig:HumanAnalysis}b). 
This shows how the difficulty of the verification task is directly linked to the IoU of the detection, and is reflected in the time measurements.
We also found that human verification errors strongly correlate with the area of objects: 48\% of all errors are made when objects occupy less than 10\% of the image area (fig.~\ref{fig:HumanAnalysis}c).



\begin{figure*}[t!]
\vspace{-.5cm}
  \begin{center}
  \includegraphics[width=\linewidth]{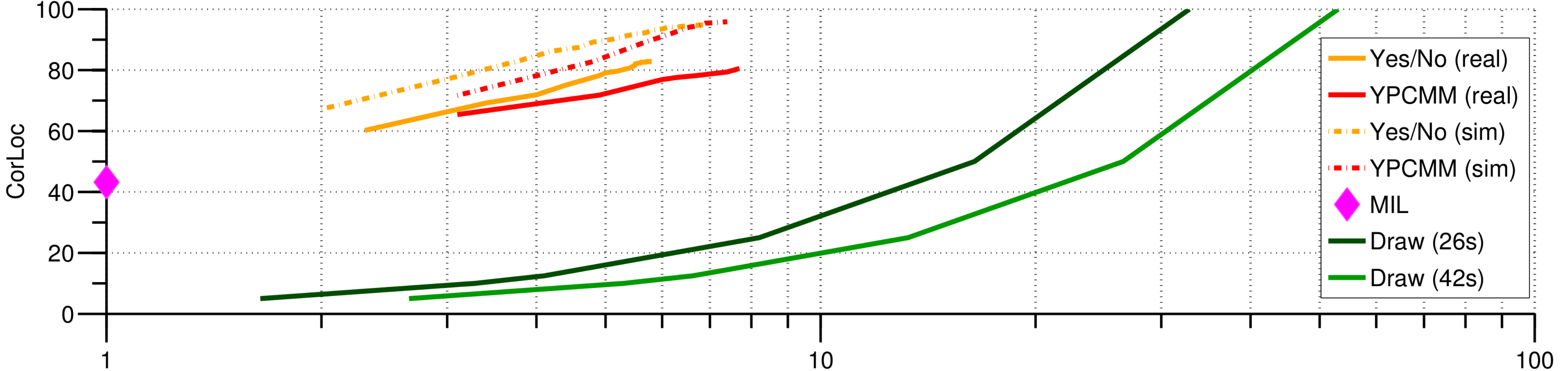}
  \includegraphics[width=\linewidth]{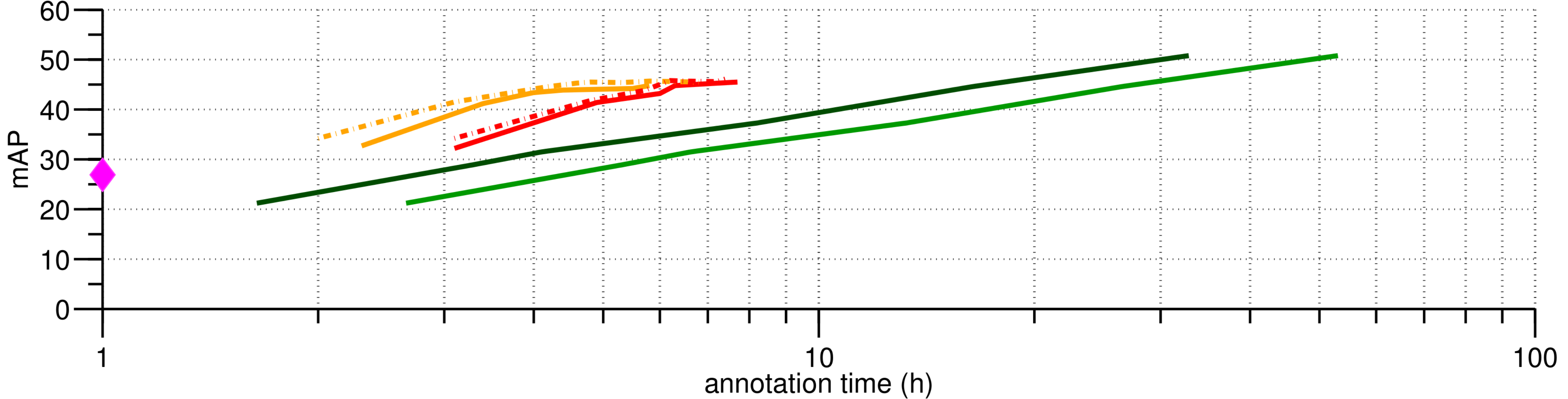}
\vspace{-7mm}
\caption{\small Evaluation on PASCAL VOC 2007: CorLoc and mAP against human
annotation time in hours (log-scale).
All orange and red curves are variations of our proposed scheme, with simulated (`sim') and real (`real') annotators.
`Draw' indicates learning from manual bounding-boxes (full supervision).
`MIL' indicates learning under weak supervision only, without human verification (sec.~\ref{cptMIL}).
The fully supervised approach needs a factor $6\times$-$9\times$ extra annotation time to obtain similar performance to our framework.
} 
\label{fig:Humans_mAP_CorLoc}
\vspace{-.6cm}
\end{center}
\end{figure*}

\mypar{Simulated vs.~human annotators.}

We first compare simulated and human
annotators by plotting CorLoc and mAP against actual annotation time (rather than as number of verifications as in Sec.~\ref{sec:pascal_sim}). For simulated verification, we use average
human annotation time as reported above. Fig.~\ref{fig:Humans_mAP_CorLoc} shows the results on a log-scale.
While human annotators are somewhat worse than simulations in terms of CorLoc on the training set,
the mAP of the resulting object detectors on the test set are comparable.
The diverging results for CorLoc and mAP is because human judgement errors are
generally made on boundary cases with bounding-boxes that only approximately cover an object (fig.~\ref{fig:HumanAnalysis}a).
Using these cases either as positive or negative training examples, the object detector remains equally strong. To conclude, in terms of training high quality object detectors, simulated human annotators reliably deliver similar results as actual annotators.

In sec.~\ref{sec:pascal_sim} we observed that YPCMM needs fewer verifications than Yes/No. However, in terms of total annotation time, the Yes/No task has the more favourable trade-off: Yes/No achieves achieves 83\% CorLoc and 45\% mAP by taking 5.8 hours of annotation time, while YPCMM achieves 81\% CorLoc and 45\% mAP by taking 7.7 hours (fig.~\ref{fig:Humans_mAP_CorLoc}). 
Hence we conclude that the Yes/No task is preferable for human annotation, as it is easier and faster.

\mypar{Weak supervision vs.~verification.}
We now compare the reference MIL WSOL approach that we use to initialize our process (Sec.~\ref{cptMIL} and magenta diamond in fig.~\ref{fig:Humans_mAP_CorLoc}) to the final output of our Yes/No human verification scheme (solid orange line, fig.~\ref{fig:Humans_mAP_CorLoc}).
While MIL WSOL achieves 43\% CorLoc and 27\% mAP, using
human verification bring a massive jump in performance to 83\% CorLoc and 45\% mAP.
Hence at a modest cost of 5.8 hours of annotation time we achieve substantial performance gains.
Examples in fig.~\ref{fig:WS_to_verification} show that our approach localizes objects more accurately and succeeds in more challenging conditions, e.g. when the object is very small and appears in a cluttered scene.

The state-of-the-art WSOL approaches perform as follows:
Cinbis et al.~\cite{cinbis15pami} achieve 52.0\% CorLoc and 30.2\% mAP,
Bilen et al.~\cite{bilen15cvpr} achieve 43.7\% CorLoc and 27.7\% mAP,
and Wang et al.~\cite{wang15tip} achieve 48.5\% CorLoc and 31.6\% mAP.
Our method using human verification substantially outperforms all of them, reaching 83\% CorLoc and 45\% mAP (our method and~\cite{bilen15cvpr,cinbis15pami,wang15tip} all use AlexNet). Hence at a modest extra annotation cost, we obtain many more correct object locations and train better detectors.

\mypar{Full supervision vs.~verification}
We now compare our Yes/No human verification scheme (solid orange line, fig.~\ref{fig:Humans_mAP_CorLoc}) to standard fully supervised learning with manual bounding-boxes (solid green lines). 
The object detectors learned by our scheme achieve 45\% mAP, almost as good as the fully supervised ones (51\% mAP).
Importantly, fully supervised training needs 33~hours of annotation time (when assuming an optimistic 26~s per image), or even 53~hours (when assuming a more realistic 42~s per image).  Our method instead requires
only 5.8 hours, a reduction in human effort of a factor of $6\times$-$9\times$.

From a different perspective, when given the same human annotation time as our approach (5.8 hours), the fully supervised detector only achieves 33\% mAP (at 26 s per bounding-box) or 30\% mAP (at 42 s).

We conclude that by the use of just an inexpensive verification component, we can train strong
object detectors at little cost. This is significant since it enables the cheap creation of high
quality object detectors for a large variety of classes, bypassing the need for massive
annotation efforts such as ImageNet~\cite{russakovsky15ijcv}.

\subsection{Complete training set and VGG16}
\label{sec:pascal_full}

\begin{table}[t]
\centering
\small
\begin{tabular}{|l|c|c|c|c|}
\hline
& \multicolumn{2}{|c|}{reduced training set} &
\multicolumn{2}{|c|}{complete training set} \\
\hline
& Yes/No & FS &  Yes/No & FS \\
\hline
AlexNet & 45\% & 51\% & 50\% & 55\% \\
VGG16 & 55\% & 61\% & 58\% & 66\% \\
\hline
\end{tabular}
\vspace{-2mm}
\caption{\small Comparison of mAP results between our Yes/No human verification scheme and full supervision (FS) using different training sets and different network architectures. `reduced training set': excluding trainval images containing only difficult and truncated instances (3550 images); `complete training set': all trainval images (5011).}
\label{tabAllMapResults}
\vspace{-5mm}
\end{table}

Our experiments are based on the Fast R-CNN detector~\cite{girshick15iccv}.
In order to have a clean comparison between our verification-based scheme and the fully supervised results of~\cite{girshick15iccv}, we re-ran our experiments using the complete trainval set of PASCAL VOC 2007 (i.e.~5011 images, table~\ref{tabAllMapResults}).
Under full supervision,~\cite{girshick15iccv} reports 57\% mAP based on AlexNet. Training Fast R-CNN from one bounding-box per class per image, results in 55\% mAP, while our Yes/No human verification scheme gets to 50\% mAP.
Additionally, we experiment with VGG16 instead of AlexNet, with the same settings. Training with full supervision leads to 66\% mAP, while our verification scheme delivers 58\% mAP.
Hence, on both CNN architectures our verification-based training scheme produces high quality detectors, 
achieving 90\% of the mAP of their fully supervised counterparts.

\subsection{Conclusions}

We proposed a scheme for training object class detectors which uses a human verification step to improve the re-training and re-localisation steps common to most weakly supervised approaches. 
Experiments on PASCAL VOC 2007 show that our scheme produces detectors performing almost as good as those trained in a fully supervised setting, {\em without ever drawing any bounding-box}. As the verification task is very quick, our scheme reduces the total human annotation time by a factor of $6\times$-$9\times$.

\mypar{Acknowledgement.} This work was supported by the ERC Starting Grant ``VisCul''.

{\small
\bibliographystyle{ieee}
\bibliography{../../../bibtex/shortstrings,../../../bibtex/vggroup,../../../bibtex/calvin}
}

\end{document}